\documentclass{article}

\usepackage[nonatbib,final]{neurips_2020}

\usepackage[utf8]{inputenc} % allow utf-8 input
\usepackage[T1]{fontenc}    % use 8-bit T1 fonts
\usepackage{url}            % simple URL typesetting
\usepackage{booktabs}       % professional-quality tables
\usepackage{amsfonts}       % blackboard math symbols
\usepackage{nicefrac}       % compact symbols for 1/2, etc.
\usepackage{adjustbox}
\usepackage{microtype}      % microtypography
\usepackage{multirow}
\usepackage{xspace}
\usepackage{mathrsfs}
\usepackage{times}
\usepackage[ruled,vlined]{algorithm2e}
\usepackage{amsmath}
\allowdisplaybreaks
\usepackage[table]{xcolor}
\usepackage{color,colortbl}
\definecolor{LightCyan}{rgb}{0.88,1,1}
\definecolor{GrayDark}{RGB}{191,191,191}
\definecolor{GrayLight}{RGB}{217,217,217}
\usepackage[colorlinks=true,linkcolor=blue, urlcolor=magenta,citecolor=blue, anchorcolor=blue]{hyperref}

\usepackage[title]{appendix}
\usepackage{wrapfig}
\usepackage{amsthm}
\usepackage[font=small,labelfont=bf]{caption}
\newtheorem{theorem}{Theorem}

\providecommand{\e}[1]{\ensuremath{\times 10^{#1}}}
\newcommand{\ie}{\textit{i.e.}}
\newcommand{\eg}{\textit{e.g.}}
\newcommand{\etal}{\textit{et al.}\xspace}
\newcommand{\cut}[1]{}
\newcommand{\xhdr}[1]{\noindent{{\bf #1.}}}

\newcommand{\std}[1]{\scriptsize{$\pm$#1}}

\usepackage{bbding}
\usepackage{pifont}
\definecolor{darkpastelgreen}{rgb}{0.01, 0.75, 0.24}
\newcommand{\greencheck}{{\color{darkpastelgreen}\CheckmarkBold}}
\newcommand{\redmark}{{\color{red}\ding{55}}}

\SetCommentSty{mycommfont}

\newcommand{\mname}{\textsc{G-Meta}\xspace}
\newcommand{\name}{\textsc{G-Meta}\xspace}
\newcommand{\longname}{\textsc{G-Meta}\xspace}

\usepackage{contour}
\usepackage{ulem}

\contourlength{0.8pt}
\newcommand{\myuline}[1]{%
  \uline{\phantom{#1}}%
  \llap{\contour{white}{#1}}%
}
\renewcommand{\underline}{\myuline}

\title{Graph Meta Learning via Local Subgraphs}
\author{%
  Kexin Huang \\
  Harvard University\\
  \texttt{kexinhuang@hsph.harvard.edu} \\
  \And
  Marinka Zitnik \\
  Harvard University\\
  \texttt{marinka@hms.harvard.edu}
}

\begin{document}

\maketitle

\begin{abstract}
Prevailing methods for graphs require abundant label and edge information for learning. When data for a new task are scarce, meta learning can learn from prior experiences and form much-needed inductive biases for fast adaption to new tasks. 
Here, we introduce \name, a novel meta-learning algorithm for graphs. 
\name uses local subgraphs to transfer subgraph-specific information and learn transferable knowledge faster via meta gradients. \name learns how to quickly adapt to a new task using only a handful of nodes or edges in the new task and does so by learning from data points in other graphs or related, albeit disjoint, label sets. 
\name is theoretically justified as we show that the evidence for a prediction can be found in the local subgraph surrounding the target node or edge.
Experiments on seven datasets and nine baseline methods show that \name outperforms existing methods by up to 16.3\%. Unlike previous methods, \name successfully learns in challenging, few-shot learning settings that require generalization to completely new graphs and never-before-seen labels. Finally, \name scales to large graphs, which we demonstrate on a new Tree-of-Life dataset comprising 1,840 graphs, a two-orders of magnitude increase in the number of graphs used in prior work. 
\end{abstract}

% Theoretical Analysis on the subgraph network effect of node classification. 

% how to define transferability 

\section{Introduction}\label{sec:intro}

Graph Neural Networks (GNNs) have achieved remarkable results in domains such as recommender systems~\cite{ying2018graph}, molecular biology~\cite{zitnik2018modeling,huang2020skipgnn}, and knowledge graphs~\cite{wang2019heterogeneous,hu2020heterogeneous}. Performance is typically evaluated after extensive training on datasets where majority of labels are available~\cite{pmlr-v97-wu19e,xu2018how}. In contrast, many problems require rapid learning from only a few labeled nodes or edges in the graph. Such flexible adaptation, known as meta learning, has been extensively studied for images and language, \eg, \cite{sung2018learning,wortsman2019learning,mccann2018natural}. However, meta learning on graphs has received considerably less research attention and has remained a problem beyond the reach of prevailing GNN models. 

\begin{figure}
\centering
\includegraphics[width=\textwidth]{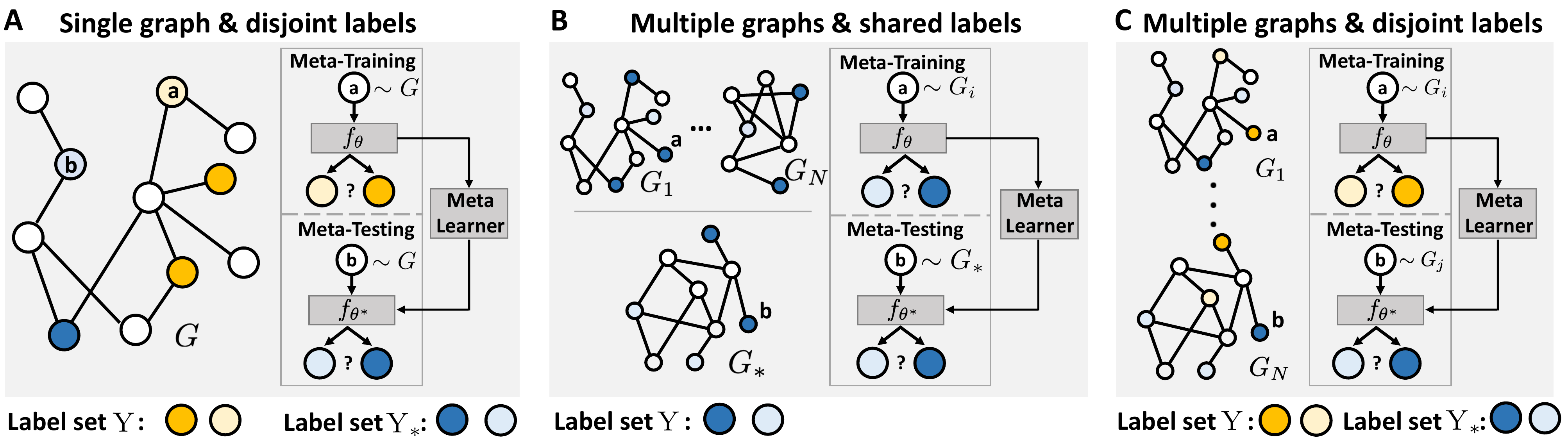}
\caption{Graph meta-learning problems. \textbf{A.} Meta-learner classifies unseen label set by observing other label sets in the same graph. \textbf{B.} Meta-learner learns unseen graph by learning from other graphs with the same label set. \textbf{C.} Meta-learner classifies unseen label set by learning from other label sets across multiple graphs. Unlike existing methods, \name solves all three problems and also works for link prediction (see Section~\ref{sec:results}).}
\vspace{-3mm}
\label{fig:conceptual}
\end{figure}

Meta learning on graphs generally refers to a scenario in which a model learns at two levels. In the first level, rapid learning occurs \textit{within} a task. For example, when a GNN learns to classify nodes in a particular graph accurately. In the second level, this learning is guided by knowledge accumulated gradually \textit{across} tasks to capture how the task structure changes across target domains~\cite{schmidhuber1997shifting,caruana1997multitask,santoro2016meta}. A powerful GNN trained to meta-learn can quickly learn never-before-seen labels and relations using only a handful of labeled data points. As an example, a key problem in biology is to translate insights from non-human organisms (such as yeast, zebrafish, and mouse) to humans~\cite{zitnik2019evolution}. How to train a GNN to effectively meta-learn on a large number of incomplete and scarcely labeled protein-protein interaction (PPI) graphs from various organisms, transfer the accrued knowledge to humans, and use it to predict the roles of protein nodes in the human PPI graph? While this is a hard task, it is only an instance of a particular graph meta-learning problem (Figure~\ref{fig:conceptual}C). In this problem, the GNN needs to learn on a large number of graphs, each scarcely labeled with a unique label set. Then, it needs to quickly adapt to a never-before-seen graph (\eg, a PPI graph from a new organism) and never-before-seen labels (\eg, newly discovered roles of proteins). Current methods are specialized techniques specifically designed for a particular problem and a particular task~\cite{zhou2019meta,bose2019meta,NIPS2019_8389,chen-etal-2019-meta,xiong-etal-2018-one}. While these methods provide a promising approach to meta learning in GNNs, their specific strategy does not scale well nor extend to other problems (Figure~\ref{fig:conceptual}).

\xhdr{Present work}
We introduce \name,\footnote{Code and datasets are available at \url{https://github.com/mims-harvard/G-Meta}.} an approach for meta learning on graphs (Figure~\ref{fig:conceptual}). The core principle of \name is to represent every node with a local subgraph and use subgraphs to train GNNs to meta-learn.
Our theoretical analysis (Section~\ref{sec:theory}) suggests that the evidence for a prediction can be found in the subgraph surrounding the target node or edge when using GNNs. In contrast to \name, earlier techniques are trained to meta-learn on entire graphs. As we show theoretically and empirically, such methods are unlikely to succeed in few-shot learning settings when the labels are scarce and scattered around multiple graphs. 
Furthermore, previous methods capture the overall graph structure but at the loss of finer local structures. Besides, \name's construction of local subgraphs gives local structural representations that enable direct structural similarity comparison using GNNs based on its connection to Weisfeiler-Lehman test~\cite{xu2018how,zhang2017weisfeiler}. Further, structural similarity enables \name to form the much-needed inductive bias via a metric-learning algorithm~\cite{snell2017prototypical}. Moreover, local subgraphs also allow for effective feature propagation and label smoothing within a GNN. 

(1) \name is a general approach for a variety of meta learning problems on graph-structured data. While previous  methods~\cite{zhou2019meta,bose2019meta} apply only to one graph meta-learning problem~(Figure~\ref{fig:conceptual}), \name works for all of them (Appendix~\ref{appendix:problem}). 
(2) \name yields accurate predictors. We demonstrate \name's performance on seven datasets and compare it against nine baselines. \name considerably outperforms baselines by up to 16.3\%. 
(3) \name is scalable. By operating on subgraphs, \name needs to examine only small graph neighborhoods. We show how \name scales to large graphs by applying it to our new Tree-of-Life dataset comprising 1,840  graphs, a two-orders of magnitude increase in the number of graphs used in prior work.

\section{Related Work}\label{sec:related}

(1) \underline{Few-shot meta learning}. Few-shot meta learning learns from prior experiences and transfers them to a new task using only a few labeled examples~\cite{vilalta2002perspective,thrun2012learning}. Meta learning methods generally fall into three categories, model-based~\cite{graves2014neural,weston2014memory,santoro2016meta}, metric-based~\cite{snell2017prototypical,8578229,NIPS2016_6385}, and optimization-based~\cite{Ravi2017OptimizationAA,nichol2018first,finn2017model} techniques. 
(2) \underline{Meta learning for graphs}. Recent studies incorporate graph-structured data into meta learning. Liu \etal~\cite{NIPS2019_8389} construct image category graphs to improve few-shot learning. Zhou \etal~\cite{zhou2020fast} use graph meta learning for fast network alignment. MetaR~\cite{chen-etal-2019-meta} and GMatching~\cite{xiong-etal-2018-one} use metric methods to generalize over new relations from a handful of associative relations in a knowledge graph. While these methods learn from a single graph, \name can handle settings with many graphs and disjoint label sets. Further, Chauhan \etal~\cite{Chauhan2020FEW-SHOT} use a super-class prototypical network for graph classification. In contrast, \name focuses on node classification and link prediction. In settings with single graphs and disjoint labels (Figure~\ref{fig:conceptual}), Meta-GNN~\cite{zhou2019meta} uses gradient-based meta-learning for node classification. On a related note, GFL~\cite{yao2019graph} focuses on settings with multiple graphs and shared labels across the graphs. Similarly, Meta-Graph~\cite{bose2019meta} uses graph signature functions for few-shot link prediction across multiple graphs. In contrast, our \name can be used for many more problems (Figure~\ref{fig:conceptual}). Further, Meta-GNN operates a GNN on an entire graph, whereas \name extracts relevant local subgraphs first and then trains a GNN on each subgraph individually. In Meta-GNN, a task is defined as a batch of node embeddings, whereas in \name, a task is given by a batch of subgraphs. This difference allows for rapid adaptation to new tasks, improves scalability, and applies broadly to meta learning problems. In summary, \name works for all problems in Figure~\ref{fig:conceptual} and applies to node classification and link prediction.
(3) \underline{Subgraphs and GNNs}. The ability to model subgraph structure is vital for numerous graph tasks, \eg, \cite{infomax,donnat2018learning,ying2019gnnexplainer,teru2019inductive}. For example, Patchy-San~\cite{patchysan} uses local receptive fields to extract useful features from graphs. Ego-CNN uses ego graphs to find critical graph structures~\cite{egocnn}. SEAL~\cite{zhang2018link} develops theory showing that enclosing subgraphs capture graph heuristics, which we extend to GNNs here. Cluster-GCN~\cite{chiang2019cluster} and GraphSAINT~\cite{Zeng2020GraphSAINT:} use subgraphs to improve GNN scalability. \name is the first approach to use subgraphs for meta learning. 

\section{Background and Problem Formulation}\label{sec:formulation}

Let $\mathcal{G} = \{G_1, \dots, G_N \}$ denote $N$ graphs. For each graph $G = (\mathcal{V}, \mathcal{E}, \mathbf{X})$, $\mathcal{V}$ is a set of nodes, $\mathcal{E}$ is a set of edges, and $\mathbf{X} = \{\mathbf{x}_1, \dots, \mathbf{x}_n\}$ is a set of attribute vectors, where $\mathbf{x}_u \in \mathbb{R}^d$ is a $d$-dimensional feature vector for node $u \in \mathcal{V}$. We denote $\mathcal{Y} = \{Y_1, \dots, Y_M\}$ as a set of $M$ distinct labels. We use $\textsc{Y}$ to denote a set of labels selected from $\mathcal{Y}$.
\name's core principle is to represent nodes with local subgraphs and then use subgraphs to transfer knowledge across tasks, graphs, and sets of labels. We use $\mathcal{S}$ to denote local subgraphs for nodes, $\mathcal{S} = \{S_1, \dots, S_n \}$. Node classification aims to specify a GNN $f_\theta: \mathcal{S} \mapsto \{1, \dots, |\textsc{Y}|\}$ that can accurately map node $u$'s local subgraph $S_u$ to labels in $\textsc{Y}$ given only a handful of labeled nodes. 

\xhdr{Background on graph neural networks} GNNs learn compact representations (embeddings) that capture network structure and node features. A GNN generates outputs through a series of propagation layers~\cite{gilmer2017neural}, where propagation at layer $l$ consists of the following three steps: (1) \underline{Neural message passing}. GNN computes a message $\mathbf{m}^{(l)}_{uv} = \textsc{Msg}(\mathbf{h}_u^{(l-1)}, \mathbf{h}_v^{(l-1)})$ for every linked nodes $u,v$ based on their embeddings from the previous layer $\mathbf{h}_u^{(l-1)}$ and $\mathbf{h}_v^{(l-1)}$. (2)~\underline{Neighborhood aggregation}. The messages between node $u$ and its neighbors $\mathcal{N}_u$ are aggregated as $\hat{\mathbf{m}}^{(l)}_{u} = \textsc{Agg}({\mathbf{m}^{(l)}_{uv} | v \in \mathcal{N}_u})$. (3) \underline{Update}. Finally, GNN uses a non-linear function to update node embeddings as $\mathbf{h}^{(l)}_u = \textsc{Upd}(\hat{\mathbf{m}}^{(l)}_{u}, \mathbf{h}^{(l-1)}_u)$ using the aggregated message and the embedding from the previous layer.

\xhdr{Background on meta learning} In meta learning, we have a meta-set $\mathscr{D}$ that consists of $\mathscr{D}_\text{train}, \mathscr{D}_\text{val}, \mathscr{D}_\text{test}$. This meta-set consists of many tasks. Each task $\mathcal{T}_i \in \mathscr{D}$ can be divided into $\mathcal{T}_i^\text{support}$ and $\mathcal{T}_i^\text{query}$. $\mathcal{T}_i^\text{support}$ has $\textsc{k}_\text{support}$ labeled data points in each label for learning and $\mathcal{T}_i^\text{query}$ has $\textsc{k}_\text{query}$ data points in each label for evaluation. The size of the label set for the meta-set $\vert \textsc{Y} \vert$ is $\textsc{N}$. It is also called $\textsc{N}$-ways $\textsc{k}_\text{support}$-shots learning problem. During meta-training, for $\mathcal{T}_i$ from $\mathscr{D}_\text{train}$ , the model first learns from $\mathcal{T}_i^\text{support}$ and then evaluates on $\mathcal{T}_i^\text{query}$ to see how well the model performs on that task. The goal of Model-Agnostic Meta-Learning (MAML)~\cite{finn2017model} is to obtain a parameter initialization $\theta_*$ that can adapt to unseen tasks quickly, such as $\mathscr{D}_\text{test}$, using gradients information learnt during meta-training.  Hyperparameters are tuned via $\mathscr{D}_\text{val}$.

\subsection{\name: Problem Formulation} \label{sec: task_setup}

\name is designed for three fundamentally different meta-learning problems~(Figure~\ref{fig:conceptual}). ``Shared labels'' refers to the situation where every task shares the same label set $\textsc{Y}$. ``Disjoint labels'' refers to the situation where label sets for tasks $i$ and $j$ are disjoint, \ie, $\textsc{Y}_i \cap \textsc{Y}_j = \emptyset$. Each data point in a task $\mathcal{T}_i$ is a local subgraph $S_u$, along with its associated label $Y_u$. \name aims to adapt to a new task $\mathcal{T}_* \sim p(\mathcal{T})$ for which only a handful examples are available after observing related tasks $\mathcal{T}_i \sim p(\mathcal{T})$. 

\xhdr{Graph meta-learning problem 1: Single Graph and Disjoint Labels} We have graph $G$ and a distribution of label sets $p(\textsc{Y}|G)$. The goal is to adapt to an unseen label set $\textsc{Y}_* \sim p(\textsc{Y}|G)$ by learning from tasks with other label sets $\textsc{Y}_i \sim p(\textsc{Y}|G)$, where $\textsc{Y}_i \cap \textsc{Y}_* = \emptyset$ for every label set $\textsc{Y}_i$.   
\xhdr{Graph meta-learning problem 2: Multiple Graphs and Shared Labels}
We have a distribution of graphs $p(G)$ and one label set $\textsc{Y}$. The goal is to learn from graph $G_j \sim p(G)$ and quickly adapt to an unseen graph $G_* \sim p(G)$, where $G_j$ and $G_*$ are disjoint. All tasks share the same labels. 

\xhdr{Graph meta-learning problem 3: Multiple Graphs and Disjoint Labels} We have a distribution of label sets $p(\textsc{Y}|\mathcal{G})$ conditioned on multiple graphs $\mathcal{G}$. Each task has its own label set $\textsc{Y}_i$ but the same label set can appear in multiple graphs. The goal is to adapt to an unseen label set $\textsc{Y}_* \sim p(\textsc{Y}|\mathcal{G})$ by learning from a disjoint label set $\textsc{Y}_i \sim p(\textsc{Y}|\mathcal{G})$, where $\textsc{Y}_i \cap \textsc{Y}_* = \emptyset$.

\section{Local Subgraphs and Theoretical Motivation for \name}\label{sec:theory}

We start by describing how to construct local subgraphs in \name. We then provide a theoretical justification showing that local subgraphs preserve useful information from the entire graph. We then argue how subgraphs enable \name to capture sufficient information on the graph structure, node features, and labels, and use that information for graph meta-learning.

For node $u$, a local subgraph is defined as a subgraph $S_u = (\mathcal{V}^u, \mathcal{E}^u, \mathbf{X}^u)$ induced from a set of nodes $\{v | d(u, v) \le h\}$, where $d(u, v)$ is the shortest path distance between node $u$ and $v$, and $h$ defines the neighborhood size. In a meta-task, subgraphs are sampled from graphs or label sets, depending on the graph meta-learning problems defined in Section~\ref{sec:formulation}. We then use GNNs to encode the local subgraphs. However, one straightforward question raised is if this subgraph loses information by excluding nodes outside of it. Here, we show in theory that applying GNN on the local subgraph preserve useful information compared to using GNN on the entire graph. 

\xhdr{Preliminaries and definitions}
Next, we use GCN~\cite{kipf2017semi} as an exemplar GNN to understand how nodes influence each other during neural message passing. The assumptions are based on~\cite{wang2020unifying} and are detailed in Appendix~\ref{appendix:1}. We need the following definitions.
(1) \underline{Node influence} $ I_{u,v}$ of $v$ on $u$ in the final GNN output is: $ I_{u,v} = \Vert \partial \mathbf{x}_u^{(\infty)} / \partial \mathbf{x}_v^{(\infty)} \Vert$, where the norm is any subordinate norm and the Jacobian measures how a change in $v$ translates to a change in $u$~\cite{wang2020unifying}. 
(2) \underline{Graph influence} $I_G$ on $u$ is: $I_G(u) = \Vert [I_{u,v_1}, \dots, I_{u,v_n}] \Vert_1$, where $[I_{u,v_1}, \dots, I_{u,v_n}]$ is a vector representing the influence of other nodes on $u$. 
(3) \underline{Graph influence loss} $R_h$ is defined as: $R_h(u) = I_G(u) - I_{S_u}(u)$, where $I_G(u)$ is the influence of entire graph $G$, and $I_{S_u}(u)$ is the influence of local subgraph $S_u$. 
Next, we show how influence spreads between nodes depending on how far the nodes are from each other in a graph. 

\begin{theorem}[\textbf{Decaying Property of Node Influence}] \label{thm:node-influence}
Let $t$ be a path between node $u$ and node $v$ and let $D_\mathrm{GM}^t$ be a geometric mean of node degrees occurring on path $t$. Let $D^{t_*}_\mathrm{GM} = \min_t \{D_\mathrm{GM}^t\}$ and $h_* = d(u,v)$. Consider the node influence $I_{u,v}$ from $v$ to $u$. Then, $I_{u,v} \le C / (D^{t_*}_{\mathrm{GM}})^{h_*}. $
\end{theorem}
The proof is deferred to Appendix~\ref{appendix:1}.
Theorem~\ref{thm:node-influence} states that the influence of a node $v$ on node $u$ decays exponentially as their distance $h_*$ increases. A side theoretical finding is that the influence is significantly decided by the accumulated node degrees of the paths between two nodes. In other words, if the paths are straight lines of nodes (low accumulated node degrees), then the node influence is high. Otherwise, if the paths consist of numerous connections to other nodes (high accumulated node degrees), the node influence is minimum. Intuitively, high accumulated node degrees can bring complicated messages along the paths, which dampen each individual node influence whereas low degrees paths can pass the message directly to the target node. Since real-world graphs are usually complicated graphs with relatively high node degrees, the node influence will be considerably low. 
We then show that for a node $u$, operating GNN on a local subgraph does not lose useful information than operating GNN on the entire graph. 
\begin{theorem}[\textbf{Local Subgraph Preservation Property}]\label{thm:subgraph-preservation} Let $S_u$ be a local subgraph for node $u$ with neighborhood size $h$. Let node $v$ be defined as: $v = \mathrm{argmax}_w(\{I_{u,w} | w \in \mathcal{V} \setminus \mathcal{V}^u\})$. Let $\bar{t}$ be a path between $u$ and $v$ and let $D_{\mathrm{GM}}^{\bar{t}}$ be a geometric mean of node degrees occurring on path $\bar{t}$. Let $D^{\bar{t}_*}_\mathrm{GM} = \min_{\bar{t}}\{D_\mathrm{GM}^{\bar{t}}\}$. The following holds: $R_h(u) \le C / (D^{\bar{t}_*}_\mathrm{GM})^{h+1}.$ 
\end{theorem}
The proof is deferred to Appendix~\ref{appendix:2}. Theorem~\ref{thm:subgraph-preservation} states that the graph influence loss is bounded by an exponentially decaying term as $h$ increases. In other words, the local subgraph formulation is an $h$-th order approximation of applying GNN for the entire graph.

\xhdr{Local subgraphs enable few-shot meta-learning} 
Equipped with the theoretical justification, we now describe how the local subgraph allows \name to do few-shot meta-learning. The construction of local subgraphs captures the following. 
(1) \underline{Structures}. Graph structures present an alternative source of strong signal for prediction~\cite{donnat2018learning,infomax}, especially when node labels are scarce. The GNN representations cannot fully capture large graphs structure because they are too complicated~\cite{bose2019meta,xu2018how}. However, they can learn representations that capture the structure of small graphs, such as our local subgraphs, as is evidenced by the connection to the Weisfeiler-Lehman test~\cite{xu2018how,zhang2017weisfeiler}. Hence, subgraphs enable \name to capture structrual node information.
(2) \underline{Features}. Local subgraphs preserve useful information, as indicated by the theorems above. 
(3) \underline{Labels}. When only a handful of nodes are labeled, it is challenging to efficiently propagate the labels through the entire graph ~\cite{zhu2003semi,li2018deeper}. Metric-learning methods~\cite{snell2017prototypical} learn a task-specific metric to classify query set data using the closest point from the support set. It has been proved as an effective inductive bias~\cite{snell2017prototypical,triantafillou2019meta}. Equipped with subgraph representations that capture both structure and feature information, \name uses metric-learning by comparing the query subgraph embedding to the support subgraph embedding. As such, it circumvents the problem of having too little label information for effective propagation.

\section{\longname: Meta Learning via Local Subgraphs}\label{sec:method}
\begin{figure}[t]
\centering
\includegraphics[width=\textwidth]{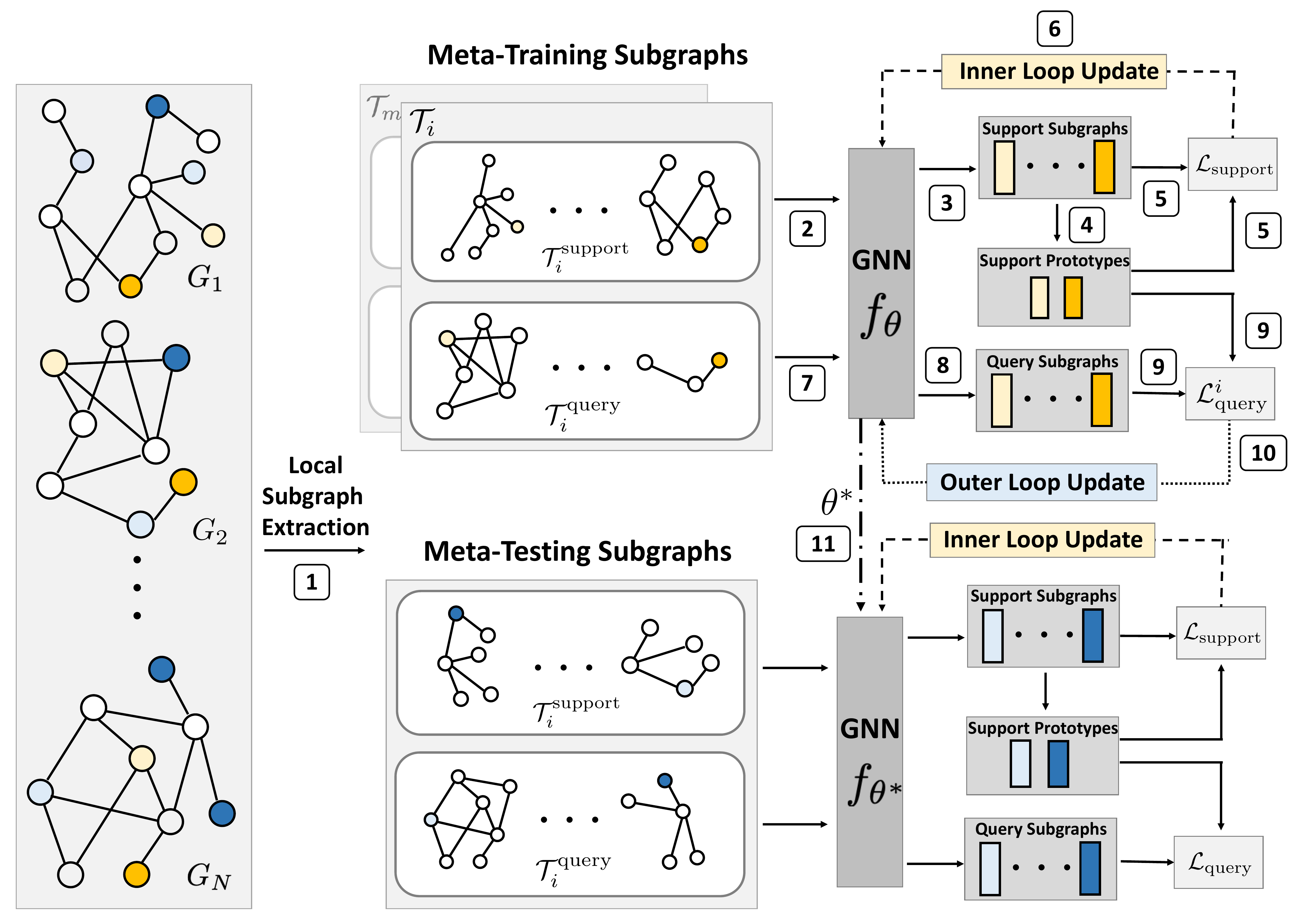}
\caption{(1) We first construct a batch of $m$ meta-training tasks and extract local subgraphs on the fly for nodes in the meta-tasks. For each task $\mathcal{T}_i$, (2) subgraphs from the support set are mini-batched and are fed into a GNN parameterized by $\theta$. (3) The support set embeddings using the centroid nodes are generated, and (4) the prototypes are computed from the support centroid embeddings. Then, (5) the support set loss $\mathcal{L}_\text{support}$ is computed, and (6) back-propagates to update the GNN parameter. (7) $\mathcal{T}^\text{query}_i$ subgraphs then feed into the updated GNN to (8) generate query centroid embeddings. (9) Using the support prototypes and the query embeddings, the query loss $\mathcal{L}_\text{query}^i$ for task $\mathcal{T}_i$ is computed. Steps (2-9) are repeated for $\eta$ update steps. The same process repeats for the other $m$ sampled tasks, starting from the same GNN $f_\theta$. (10) The last update step's query loss from all the tasks are summed up and used to update $\theta$. Then, another batch of tasks are sampled, and step (1-10) are repeated. Then, for meta-testing tasks, steps (1-9) are repeated with the GNN using the meta-learned parameter $\theta_*$, which enables generalization over unseen tasks. See Algorithm~\ref{algo1}~(Appendix~\ref{appendix:algo}).}
\label{fig:method-overview}
\end{figure}

\name~(Figure~\ref{fig:method-overview}) is an approach for meta-learning on graphs. Building on theoretical motivation from Section~\ref{sec:theory}, \name first constructs local subgraphs. It then uses a GNN encoder to generate embeddings for subgraphs. Finally, it uses prototypical loss for inductive bias and MAML for knowledge transfer across graphs and labels. The overview is in Algorithm~\ref{algo1}~(Appendix~\ref{appendix:algo}). 

\xhdr{Neural encoding of subgraphs}
In each meta-task, we first construct a subgraph $S_u$ for each node $u$. While we use $h$-hops neighbors to construct subgraphs, other subgraph extraction algorithms, e.g.,~\cite{chen2010dense,faust2010pathway} can be considered. We then feed each subgraph $S_u$ into a $h$-layer GNN to obtain an embedding for every node in the subgraph. Here, $h$ is set to the size of subgraph neighborhood. The centroid node $u$'s embedding is used to represent the subgraph $\mathbf{h}_u = \mathrm{Centroid}(\text{GNN}(S_u))$. Note that a centroid node embedding is a particular instantiation of our framework. One can consider alternative subgraph representations, such as subgraph neural networks~\cite{alsentzer2020subgraph,lou2020neural} or readout functions specified over nodes in a subgraph~\cite{xu2018how,Chauhan2020FEW-SHOT}. Notably, local subgraphs in our study are different from computation graphs~\cite{ying2019gnnexplainer}. Local subgraphs are not used for neural message passing; instead we use them for meta learning. Further, our framework does not constrain subgraphs to $h$-hop neighborhoods or subgraph neural encodings to centroid embeddings.

%Note that for faster computation, we mini-batch the subgraphs for each meta-task's support and query set.
%Without constructing local subgraphs, it is unable to do many graph meta learning problems.

\xhdr{Prototypical loss} 
After we obtain subgraph representations, we leverage inductive bias between representations and labels to circumvent the issue of limited label information in few-shot settings. For each label $k$, we take the mean over support set subgraph embeddings to obtain a prototype $\mathbf{c}_k$ as: $\mathbf{c}_k = 1/N_k\sum_{y_j = k} \mathbf{h}_j$. The prototype $\mathbf{c}_k$ serves as a landmark for label $k$. Then, for each local subgraph $S_u$ that exists in either support or query set, a class distribution vector $\mathbf{p}$ is calculated via the Euclidean distance between support prototypes for each class and centroid embeddings as: $\mathbf{p}_k = (\mathrm{exp}\left( - \Vert \mathbf{h}_u  - \mathbf{c}_k \Vert \right))/ (\sum_{\hat{k}} \mathrm{exp}\left( - \Vert \mathbf{h}_u  - \mathbf{c}_{\hat{k}} \Vert \right))$. Finally, we use class distribution vectors from local subgraphs to optimize a cross-entropy loss as follows: $\mathrm{L}(\mathbf{p}, \mathbf{y}) = \sum_j \mathbf{y}_j \; \mathrm{log} \; \mathbf{p}_j$, where $\mathbf{y}$ indicates a true label's one-hot encoding.

\xhdr{Optimization-based meta-learning}
To transfer the structural knowledge across graphs and labels, we use MAML, an optimization-based meta-learning approach. We break the node's dependency on a graph by framing nodes into independent local subgraph classification tasks. It allows direct adaptation to MAML since individual subgraphs can be considered as an individual image in the classic few-shot meta-learning setups. More specifically, we first sample a batch of tasks, where each task consists of a set of subgraphs. During meta-training inner loop, we perform the regular stochastic gradient descent on the support loss for each task $\mathcal{T}_i$: $\theta_j = \theta_{j-1} - \alpha \nabla \mathcal{L}_\text{support}$. The updated parameter is then evaluated using the query set, and the query loss for task $i$ is recorded as $\mathcal{L}_\text{query}^i$. The above steps are repeated for $\eta$ update steps. Then, the $\mathcal{L}_\text{query}^i$ from the last update step is summed up across the batch of tasks, and then we perform a meta-update step: $\theta = \theta - \beta \nabla \sum_i \mathcal{L}_\text{query}^i$. Next, new tasks batch are sampled, and the same iteration is applied on the meta-updated $\theta$. During meta-testing, the same procedure above is applied using the final meta-updated parameter $\theta_*$. $\theta_*$ is learned from knowledge across meta-training tasks and is the optimal parameter to adapt to unseen tasks quickly.

\xhdr{Attractive properties of \mname} 
(1)~\underline{Scalability:} \name operates on a mini-batch of local subgraphs where the subgraph size and the batch size (few-shots) are both small. This allows fast computation and low memory requirement because \name's aggregation field is smaller than that of previous methods that operate on entire graphs~\cite{bose2019meta,zhou2019meta}. We further increase scalability by sub-sampling subgraphs with more than 1,000 nodes. Also note that graph few-shot learning does not evaluate all the samples in the graph since few-shot learning means majority of labels do not exist. Thus, each mini-batch consists only of a few samples with labels (e.g., 9 in 3-way 3-shot learning), which are quick operations.
(2)~\underline{Inductive learning:} Since \name operates on different subgraphs in each GNN encoding, it forces inductiveness over unseen subgraphs. This inductiveness is crucial for few-shot learning, where the trained model needs to adapt to unseen nodes. Inductive learning also allows for knowledge transfer from meta-training subgraphs to meta-testing subgraphs. (3)~\underline{Over-smoothing regularization:} One limitation of GNN is that connected nodes become increasingly similar after multiple iterations of propagation on the same graph. In contrast, each iteration for \name consists of a batch of different subgraphs with various structures, sizes and nodes, where each subgraph is fed into the GNN individually. This prevents GNN from over-smoothing on the structure of a single graph. (4)~\underline{Few-shot learning:} \name needs only a tiny number of labeled nodes for successful learning, as demonstrated in experiments. This property is in contrast with prevailing GNNs, which require a large fraction of labeled nodes to propagate neural messages in the graph successfully. (5)~\underline{Broad applicability:} \name applies to many graph meta-learning problems (Figure~\ref{fig:conceptual}) whereas previous methods apply to at most one~\cite{bose2019meta,zhou2019meta}. Unlike earlier methods, \name works for node classification and few-shot link prediction (\ie, via local subgraphs for a pair of nodes).

\section{Experiments}\label{sec:experiments}
\begin{table}
    \centering
    \footnotesize
    \def\arraystretch{0.80}
    \caption{Dataset statistics. Fold-PPI and Tree-of-Life are new datasets introduced in this study.}
    \begin{tabular}{p{3.3cm}|c|ccccc}
    \toprule
    Dataset & Task & \# Graphs & \# Nodes & \# Edges & \# Features & \# Labels\\ \midrule
    Synthetic Cycle & Node & 10 & 11,476  & 19,687 & N/A & 17 \\ 
    Synthetic BA & Node & 10 & 2,000  & 7,647 & N/A & 10 \\ 
    ogbn-arxiv   & Node & 1 & 169,343 &1,166,243	& 128 & 40 \\
    Tissue-PPI & Node & 24 & 51,194 & 1,350,412 & 50 & 10 \\
    FirstMM-DB & Link & 41 & 56,468 & 126,024 & 5 & 2 \\ \midrule
    Fold-PPI & Node & 144 & 274,606 & 3,666,563 & 512 &  29 \\ 
    Tree-of-Life & Link & 1,840 & 1,450,633 & 8,762,166 & N/A & 2 \\
    \bottomrule
    \end{tabular}
\label{tab:data}
\end{table}

\xhdr{Synthetic datasets}
We have two synthetic datasets whose labels depend on nodes' structural roles~\cite{henderson2012rolx}, which we use to confirm that \mname captures local graph structure~(Table~\ref{tab:data}). (1) \underline{Cycle}: we use a cycle basis graph and attach a distribution of shapes: House, Star, Diamond, Fan~\cite{donnat2018learning}. The label of each node is the structural role define by the shape. We also add random edges as noise. In the multiple graphs problem, each graph is with varying distribution of number of shapes. (2) \underline{BA}: to model local structural information under a more realistic homophily graph, we construct a Barab{\'a}si-Albert (BA) graph and then plant different shapes to the graph. Then, we compute the Graphlet Distribution Vector~\cite{prvzulj2007biological} for each node, which characterizes the local graph structures and then we apply spectral clustering on this vector to generate the labels. For multiple graphs problem, a varying distribution of numbers of shapes are used to plant each BA graph. Details are in Appendix~\ref{appendix:dataset}.

\xhdr{Real-world datasets and new meta-learning datasets}
We use three real world datasets for node classification and two for link prediction to evaluate \name~(Table~\ref{tab:data}). (1) \underline{ogbn-arxiv} is a CS citation network, where features are titles, and labels are the subject areas~\cite{hu2020open}. (2) \underline{Tissue-PPI} consists of 24 protein-protein interaction networks from different tissues, where features are gene signatures and labels are gene ontology functions~\cite{zitnik2017predicting,hamilton2017inductive}. (3) \underline{Fold-PPI} is a novel dataset, which we constructed for the multiple graph and disjoint label problem. It has 144 tissue PPI networks~\cite{zitnik2017predicting}, and the labels are protein structures defined in SCOP database~\cite{andreeva2020scop}. The features are conjoint triad protein descriptor~\cite{shen2007predicting}. We screen fold groups that have more than 9 unique proteins across the networks. It results in 29 unique labels. Like many real-world graphs, in Fold-PPI, the majority of the nodes do not have associated labels. This also shows the importance of graph few-shot learning. (4) \underline{FirstMM-DB}~\cite{neumann2013graph} is the standard 3D point cloud data for link prediction across graphs, which consists of 41 graphs. (5) \underline{Tree-of-Life} is a new dataset that we constructed based on 1,840 protein interaction networks, each originating from a different species~\cite{zitnik2019evolution}. Since node features are not provided, we use node degrees instead. Details are in Appendix~\ref{appendix:dataset}.

\xhdr{Experimental setup}
We follow the standard episode training for few-shot learning~\cite{triantafillou2019meta}. We refer the readers to Section~\ref{sec: task_setup} for task setups given various graph meta learning problems. Note that in order to simulate the real-world few-shot graph learning settings, we do not use the full set of labels provided in the dataset. Instead, we select a partition of it by constructing a fixed number of meta-tasks. Here, we describe the various parameters used in this study. (1) \underline{Node classification.} For disjoint label setups, we sample 5 labels for meta-testing, 5 for meta-validation, and use the rest for meta-training. In each task, 2-ways 1-shot learning with 5 gradient update steps in meta-training and 10 gradient update steps in meta-testing is used for synthetic datasets. 3-ways 3-shots learning with 10 gradient update steps in meta-training and 20 gradient update steps in meta-testing are used for real-world datasets disjoint label settings. For multiple graph shared labels setups, 10\% (10\%) of all graphs are held out for testing (validation). The remaining graphs are used for training. For fold-PPI, we use the average of ten 2-way protein function tasks. 
(2) \underline{Link prediction.} 10\% of graphs are held out for testing and another 10\% for validation. For each graph, the support set consists of 30\% edges and the query set 70\%. Negative edges are sampled randomly to match the same number of positive edges, and we follow the experiment setting from~\cite{zhang2018link} to construct the training graph. We use 16-shots for each task, \ie, using only 32 node pairs to predict links for an unseen graph. 10 gradient update steps in meta-training and 20 gradient update steps in meta-testing are used. Each experiment is repeated five times to calculate the standard deviation of the results. Hyperparameters selection and a recommended set of them are in Appendix~\ref{appendix:param}.

\xhdr{Nine baseline methods}
\underline{Meta-Graph}~\cite{bose2019meta} injects graph signature in VGAE~\cite{kipf2016variational} to do few-shot multi-graph link prediction. \underline{Meta-GNN}~\cite{zhou2019meta} applies MAML~\cite{finn2017model} to Simple Graph Convolution~(SGC)~\cite{pmlr-v97-wu19e}. \underline{Few-shot Graph Isomorphism Network (FS-GIN)}~\cite{xu2018how} applies GIN on the entire graph and retrieve the few-shot nodes to propagate loss and enable learning. Similarly, \underline{Few-shot SGC ~(FS-SGC)}~\cite{pmlr-v97-wu19e} switches GIN to SGC for GNN encoder. Note that the previous four baselines only work in a few graph meta-learning problems. We also test on different meta-learning models, using the top performing ones in~\cite{triantafillou2019meta}. We operate on subgraph level for them since it allows comparison in all problems. \underline{No-Finetune} performs training on the support set and use the trained model to classify each query example, using only meta-testing set. \underline{KNN}~\cite{dudani1976distance, triantafillou2019meta} first trains a GNN using all data in the meta-training set and it is used as an embedding function. Then, it uses the label of the voted K-closest example in the support set for each query example. \underline{Finetune}~\cite{triantafillou2019meta} uses the embedding function generated from meta-training set and the models are then finetuned on the meta-testing set. \underline{ProtoNet}~\cite{snell2017prototypical} applies prototypical learning on each subgraph embeddings, following the standard few-shot learning setups. \underline{MAML}~\cite{finn2017model} switches ProtoNet to MAML as the meta-learner. Note that the baselines ProtoNet and MAML can be considered as an ablation of \name removing MAML and Prototypical loss respectively. 
All experiments use the same number of query points per label. We use multi-class accuracy metric. 
%\section{Results}\label{sec:results}
\begin{figure}
\begin{minipage}{\textwidth}
  \begin{minipage}[b]{0.69\textwidth}
    \centering
    \captionof{table}{\textbf{Graph meta-learning performance on synthetic datasets.} Reported is multi-class classification accuracy (five-fold average) for 1-shot node classification in various graph meta-learning problems. N/A means the method does not apply for the problem. Disjoint label problems use a 2-way setup. In the shared label problem, the cycle graph has 17 labels and the BA graph has 10 labels. The results use 5 gradient update steps in meta-training and 10 gradient update steps in meta-testing. Full table with standard deviations is in Appendix~\ref{appendix:syn}. }
    \begin{adjustbox}{width = \textwidth}
    \begin{tabular}[b]{l|cc|cc|cc}
    \toprule
    \rowcolor{GrayDark} Graph Meta- & \multicolumn{2}{c|}{Single graph} & \multicolumn{2}{c|}{Multiple graphs} & \multicolumn{2}{c}{Multiple graphs} \\ 
    \rowcolor{GrayDark} Learning Problem & \multicolumn{2}{c|}{Disjoint labels} & \multicolumn{2}{c|}{Shared labels} & \multicolumn{2}{c}{Disjoint labels} \\ \midrule 
    \rowcolor{GrayLight}Prediction Task &\multicolumn{2}{c|}{Node} & \multicolumn{2}{c|}{Node} &\multicolumn{2}{c}{Node} \\ \midrule
    Base graph & Cycle & BA & Cycle & BA & Cycle & BA \\ \midrule
    \name~(Ours) &0.872& \textbf{0.867}& \textbf{0.542}&\textbf{0.734}&\textbf{0.767}&\textbf{0.867} \\
    Meta-Graph  & N/A & N/A & N/A & N/A &  N/A &  N/A \\
    Meta-GNN & 0.720&0.694 & N/A & N/A & N/A & N/A \\
    FS-GIN & 0.684 & 0.749 & N/A & N/A & N/A & N/A \\
    FS-SGC &0.574 &0.715 & N/A & N/A & N/A & N/A \\\midrule
    KNN &\textbf{0.918} & 0.804 & 0.343 & 0.710 & 0.753&0.769 \\
    No-Finetune & 0.509 &0.567&0.059& 0.265 &0.592&0.577 \\
    Finetune &0.679 &0.671 & 0.385& 0.517 &0.599&0.629 \\ \midrule
    ProtoNet &0.821&0.858&0.282&0.657&0.749&0.866 \\
    MAML &0.842&0.848&0.511&0.726&0.653&0.844 \\ 
    \bottomrule
    \end{tabular}
    \end{adjustbox}
    \label{tab:res_syn}
    \end{minipage}
    \hfill
    \begin{minipage}[b]{0.29\textwidth}
    \centering
    \includegraphics[width=0.83\textwidth]{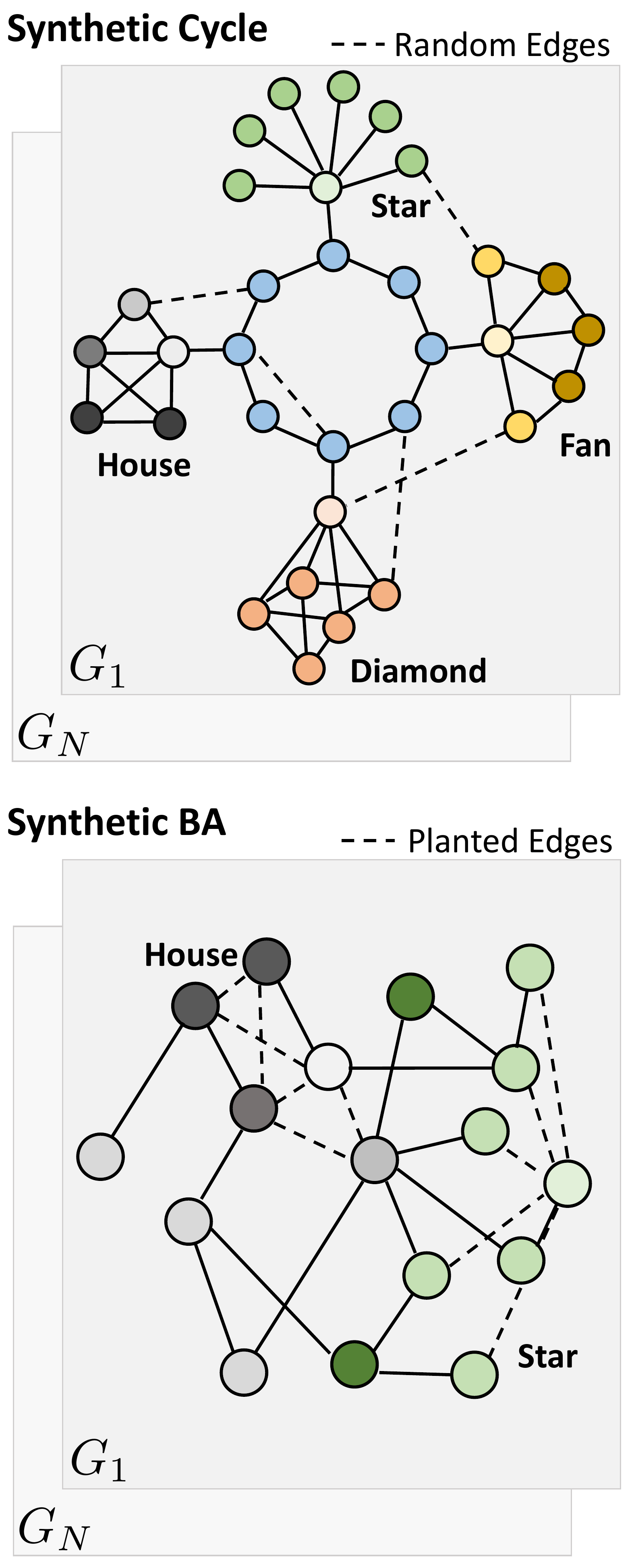}
    \captionof{figure}{Synthetic Cycle and Ba\-rab{\'a}si-Albert datasets.} \label{fig:synthetic-data-illustration}
  \end{minipage}
  \end{minipage}
  \vspace{-5mm}
\end{figure}

\begin{table}
    \centering
    \def\arraystretch{0.80}
    \caption{\textbf{Graph meta-learning performance on real-world datasets.} Reported is multi-class classification accuracy (five-fold average) and standard deviation. N/A means the method does not work in the graph meta-learning problem. Note that ProtoNet and MAML can be considered as ablation studies of \name. Further details on performance are in Appendix~\ref{appendix:perf}.}
    \begin{adjustbox}{width=\textwidth}
    \begin{tabular}{l|c|c|c|c|c}
    \toprule
    \rowcolor{GrayDark} Graph Meta- & Single graph & Multiple graphs & Multiple graphs & Multiple graphs & Multiple graphs \\ 
    \rowcolor{GrayDark} Learning Problem & Disjoint labels & Shared labels & Disjoint labels &Shared labels & Shared labels  \\\midrule 
    \rowcolor{GrayLight}Prediction Task &Node&Node&Node&Link&Link \\ \midrule
    Dataset & ogbn-arxiv & Tissue-PPI & Fold-PPI & FirstMM-DB & Tree-of-Life\\ \midrule
    \name (Ours) & \textbf{0.451}\std{0.032} & \textbf{0.768}\std{0.029} & \textbf{0.561}\std{0.059}& \textbf{0.784}\std{0.028} & \textbf{0.722}\std{0.032} \\
    Meta-Graph & N/A & N/A & N/A & 0.719\std{0.020}& 0.705\std{0.004} \\
    Meta-GNN &0.273\std{0.122} & N/A & N/A &N/A & N/A \\
    FS-GIN & 0.336\std{0.042} & N/A & N/A & N/A & N/A \\
    FS-SGC & 0.347\std{0.005} & N/A & N/A & N/A & N/A\\\midrule
    KNN & 0.392\std{0.015} & 0.619\std{0.025} & 0.433\std{0.034} &0.603\std{0.072} & 0.649\std{0.012} \\
    No-Finetune &0.364\std{0.014}& 0.516\std{0.006} & 0.376\std{0.017} &0.509\std{0.006} & 0.505\std{0.001} \\
    Finetune &0.359\std{0.010}& 0.521\std{0.013} & 0.370\std{0.022}&0.511\std{0.007} & 0.504\std{0.003} \\ \midrule
    ProtoNet & 0.372\std{0.017} & 0.546\std{0.025} & 0.382\std{0.031} & 0.779\std{0.020} & 0.697\std{0.010} \\
    MAML & 0.389\std{0.021} & 0.745\std{0.051} & 0.482\std{0.062} & 0.758\std{0.025} & 0.719\std{0.012} \\ 
    \bottomrule
    \end{tabular}
    \end{adjustbox}
    \vspace{-4mm}
    \label{tab:res_real}
\end{table}

\subsection{Results}\label{sec:results}

\xhdr{Overview of results} Synthetic dataset results are reported in Table~\ref{tab:res_syn} and real-world datasets in Table~\ref{tab:res_real}. We see that \name can consistently achieve the best accuracy in almost all tested graph meta-learning problems and on both node classification and link prediction tasks. Notably, \name achieves a 15.1\% relative increase in the single graph disjoint setting, a 16.3\% relative increase in the multiple graph disjoint label setting, over the best performing baseline. This performance boost suggests that \name can learn across graphs and disjoint labels. Besides, this increase also shows \name obtains good predictive performance, using only a few gradient update steps given a few examples on the target tasks. 

\xhdr{\name captures local structural information} In the synthetic datasets, we see that Meta-GNN, FS-GIN, and FS-SGC, which base on the entire graph, are inferior to subgraph-based methods, such as \name. This finding demonstrates that subgraph embedding can capture the local structural roles, whereas using the entire graph cannot. It further supports our theoretical motivation. In the single graph disjoint label setting, KNN achieves the best result. This result suggests that the subgraph representation learned from the trained embedding function is sufficient to capture the structural roles, further confirming the usefulness of local subgraph construction.

\xhdr{\name is highly predictive, general graph meta-learning method} Across meta-learning models, we observe \name is consistently better than others such as MAML and ProtoNet. MAML and ProtoNet have volatile results across different problems and tasks, whereas \name is stable. This result confirms our analysis of using the prototypical loss to leverage the label inductive bias and MAML to transfer knowledge across graphs and labels. By comparing \name to two relevant existing works Meta-GNN and Meta-Graph, we see \name can work across different problems. In contrast, the previous two methods are restricted by the meta graph learning problems. In real-world datasets, we observe No-Finetune is better than Finetune in many problems~(ogbn-arxiv \& Fold-PPI). This observation shows that the meta-training datasets bias the meta-testing result, suggesting the importance of meta-learning algorithm to achieve positive transfer in meta graph learning. 

\xhdr{Local subgraphs are vital for successful graph few-shot learning} We observe that Meta-GNN, FS-GIN, and FS-SGC, which operate on the entire graph, perform poorly. In contrast, \name can learn transferable signal, even when label sets are disjoint. Since the problem requires learning using only 3-shots in a large graph of around 160 thousands nodes and 1 million edges~(ogbn-arxiv), we posit that operating GNN on the entire graph would not propagate useful information. Hence, the performance increase suggests that local subgraph construction is useful, especially in large graphs. 
Finally, in the Tree-of-Life link prediction problem, we studied the largest-ever graph meta-learning dataset, spanning 1,840 PPI graphs. This experiment also supports the scalability of \name, which is enabled by the local subgraphs.

\xhdr{Ablation and parameter studies} We conduct ablations on two important components, optimization-based meta-learning and prototypical loss. Note that baseline ProtoNet and MAML are the ablations of \name. We find that both components are important to improve predictive performance. To study parameters, we first vary the size of subgraph size $h$. We find that $h = 2$ has a better performance across datasets than $h = 1$. When $h = 3$, in some dataset, it introduces noise and decreases the performance while in other one, it can improve performance since it includes more information. We suggest to use $h = 2$ since it has the most stable performance. We then vary the number of shots $k$, and observes a linear trend between $k$ and the predictive performance. The detailed quantitative results on the parameter studies are provided in the Appendix~\ref{appendix:param_studies}.

\section{Conclusion}\label{sec:conclusion}

We introduce \name, a scalable and inductive meta learning method for graphs. Unlike earlier methods, \name  excels at difficult few-shot learning tasks and can tackle a variety of meta learning problems. The core principle in \name is to use local subgraphs to identify and transfer useful information across tasks. The local subgraph approach is fundamentally different from prior studies, which use entire graphs and capture global graph structure at the loss of finer topological shapes. Local subgraphs are theoretically justified and stem from our observation that evidence for a prediction can be found in the local subgraph surrounding the target entity. \name outperforms nine baselines across seven datasets, including a new and original dataset of 1,840 graphs.  

\clearpage

\section*{Broader Impact}
Graphs represent an incredibly powerful data representation that has proved useful for numerous domains and application areas. At the same time, meta learning is a key area of machine learning research that has already demonstrated great potential for problems, such as few-shot image recognition and neural architecture search.  

\xhdr{\name advances ML research}
Our present work is at an exciting yet underexplored intersection of graph ML and meta learning, advancing the state-of-the-art and enabling practical applications of meta learning for large graph datasets. State-of-the-art GNN methods do not work when there are only a handful of labels available. Our work fills in this gap by providing a novel approach to leverage related information such as related graphs or other labels sets to aid the graph few-shot learning.  Numerous real-world graphs are only associated with scarce labels of nodes or have a large percentage of missing links. Low-resource constraint (\eg, the sparsity of graphs/scarcity of labels) is a fundamental challenge in real-world graph applications. The goal of graph meta-learning is to solve key graph ML tasks, such as node classification and link prediction, under these constraints. 

We envision numerous impactful applications of our work, as already demonstrated in the paper. We provide a brief overview of key applications for graph meta-learning, and \name in particular.

\xhdr{\name can expedite scientific discovery} Science is filled with complex systems that are modeled by graphs. The labels are usually obtained through resource-intensive experiments in laboratories. Many experiments, such as those to measure protein-protein interactions (studied in the paper), cannot yet be automated. Because of that, labels (\ie, a variety of molecular properties that need to be discovered for downstream problems like drug discovery~\cite{vamathevan2019applications}) are expensive to obtain and thus are very scarce. \name can accelerate scientific discovery by learning from related sources and quickly adapting to a new, never-before-seen task of interest given only a handful of examples. For example, \name can better annotate rare disease pathways by transferring knowledge from the mouse PPI network to the human PPI network~\cite{rolland2014proteome}. Further, PPI networks of many species are highly incomplete, even for humans, less than 30\% of all pairwise combinations of proteins have been tested for interaction so far~\cite{luck2020reference}. Graph meta-learning can help label sparse networks by learning from a handful of other networks that are well-annotated.

\xhdr{\name can bring economic values} While graphs have helped businesses make better products (\eg, \cite{ying2018graph,mcauley2015image}), many graph-structured data resources remain under-utilized because of the sparsity and scarcity of labels. \name can help tackle this problem. For example, it can improve recommendation for a new set of products that only have a few user behavior records by transferring knowledge from previous product sets in the user products recommendation networks, or help a business expand to new locations by learning from the interconnected data about the current location. 

\xhdr{\name can improve equality} \name can be used to facilitate the development of world regions by quickly learning from the developed regions. While this has previously been successfully demonstrated on satellite image data~\cite{jean2016combining}, we see many new opportunities for graph data, such as identifying what transportation lines (links) should be prioritized to add in a rural region by learning from infrastructure networks of developed regions that were similarly rural before. 

Finally, we briefly comment on potential risks of having a powerful graph meta-learning predictor. (1) Negative transfer: Meta-learning models are tricky to train and may vary across datasets and domains. It can lead to negative transfer. (2) Misuse of the methodology: Meta-learning is not universal. It works in specific settings. For example, \name is applied to few-shot settings. When labels are abundant, it might not yield considerable benefits. (3) Adversarial attacks: There are no studies on how adversarial attacks would affect meta-learning algorithms and graph algorithms in particular. We posit that in the few-shot setting, it may be particularly easy to attack the graph since each labeled example is vital for prediction.

\begin{ack}
This work is supported, in part, by NSF grant nos. IIS-2030459 and IIS-2033384, and by the Harvard Data Science Initiative. The content is solely the responsibility of the authors. We thank Chengtai Cao, Avishek Joey Bose, Xiang Zhang for helpful discussions.
\end{ack}

\bibliographystyle{plain}
\bibliography{refs}

\clearpage

\begin{appendices}
\appendix
\vspace{-4mm}
\section{Further Details on Notation} \label{sec:notation}
\begin{table}[ht]
    \centering
    \caption{\textbf{The notation used in the paper.}}
    \begin{tabular}{l|l}
    \toprule
    Notation & Description \\ \midrule
    $\mathcal{G}, G$ & A set of graphs, graph $G \in \mathcal{G}$\\
    $\mathcal{V}, \mathcal{E}, \mathbf{X}$ & A set of nodes, edges, and feature matrix for graph $G$ \\ 
    $\mathcal{Y}, \textsc{Y}, Y$ & An entire set of labels, a label set for a task, and  label $Y \in \textsc{Y}$ \\
    $\mathcal{S}, S$ & A set of subgraphs, subgraph $S \in \mathcal{S}$\\
    $u,v$ & Nodes $u$ and $v$ \\
    $i,j$ & Prediction tasks $i$ and $j$ \\ 
    $f_\theta$ & GNN model $f$ parameterized by $\theta$ \\ \midrule
    $\mathcal{D}, \mathcal{D}_\text{train}, \mathcal{D}_\text{val}, \mathcal{D}_\text{test}$ & An entire meta-set, meta-training set, meta-validation set, and meta-test set \\
    $\mathcal{T}_i, \mathcal{T}_i^\text{support}, \mathcal{T}_i^\text{query}$ & Task $i$, support set, and query set for task $i$ \\ \midrule
    $N, M$ & Number of graphs and labels \\
    $\textsc{N}, \textsc{k}_\text{support}, \textsc{k}_\text{query}$ & $\textsc{N}$-way, $\textsc{k}_\text{support}$-shot prediction for support set and $\textsc{k}_\text{query}$-shot for query set \\ \midrule
    $\eta$ & Number of steps in the inner-loop (see Algorithm~\ref{algo1}) \\ 
    $\mathcal{L}_\text{support}, \mathcal{L}_\text{query}$ & Support loss and query-set loss \\
    $\mathbf{H}$ & A matrix of centroid embeddings\\
    $\mathbf{C}$ & A matrix of prototype embeddings \\
    $\mathbf{p}$ & A class probability distribution vector \\ 
    $\mathbf{y}_\text{support}, \mathbf{y}_\text{query}$ & Support and query set label vector\\
    $\alpha, \beta$ & Learning rate of inner and outer loop update (see Algorithm~\ref{algo1}) \\ 
    \bottomrule
    \end{tabular}
    \label{tab:notation}
    \vspace{-2mm}
\end{table}

\section{Further Details on the Applicability of Methods for Meta Learning on Graphs} \label{appendix:problem}
\vspace{-2mm}

\begin{table}[h!]
    \centering    
    \caption{\textbf{A comparison of notable existing methods in the context of different graph meta-learning problems.} We see \name is able to tackle all problems whereas existing methods cannot.}
    \begin{tabular}{l|c|c|c|c}
    \toprule
     \rowcolor{GrayDark} Graph Meta- & Single graph & Single graph & Multiple graphs & Multiple graphs \\ 
     \rowcolor{GrayDark} Learning Problem & Shared labels & Disjoint labels & Shared labels & Disjoint labels \\\midrule 
    GNN, \eg, \cite{kipf2017semi,xu2018how,pmlr-v97-wu19e} & \greencheck & \redmark & \redmark & \redmark \\ 
    Meta-GNN~\cite{zhou2019meta} & \redmark & \greencheck & \redmark & \redmark \\ 
    Meta-Graph~\cite{bose2019meta} & \redmark & \redmark & \greencheck & \redmark \\ 
    \name (Ours) & \greencheck & \greencheck & \greencheck & \greencheck\\ 
    \bottomrule
    \end{tabular}
    \label{tab:setup-ours-baselines}
\end{table}

\section{Assumptions and Proof of Theorem 1}\label{appendix:1}

\subsection{Assumptions} 

We use a popular GCN model~\cite{kipf2017semi} as the exemplar GNN model. The $l$-th layer GCN propagation rule is defined as: $\mathbf{H}^{(l+1)} = \sigma (\hat{\mathbf{A}}\mathbf{H}^{(l)}\mathbf{W}^{(l)})$, where $\mathbf{H}^{(l)}, \mathbf{W}^{(l)}$ are node embedding and parameter weight matrices at layer $l$, respectively, and $\hat{\mathbf{A}} = \mathbf{D}^{-1} \mathbf{A}$ is the normalized adjacency matrix. Following~\cite{wang2020unifying}, throughout these derivations, we assume that $\sigma$ is an identity function and that $\mathbf{W}$ an identity matrix. 

\subsection{Definitions and Notation}

\textbf{Definition (Node Influence)}:\textit{
Node influence $I_{u,v}$ of $v$ on $u$ in the final GNN output is: $ I_{u,v} = \Vert \partial \mathbf{x}_u^{(\infty)} / \partial \mathbf{x}_v^{(\infty)} \Vert$, where the norm is any subordinate norm and the Jacobian measures how a change in $v$ translates to a change in $u$~\cite{wang2020unifying}. }

\textbf{Definition (Graph Influence)}:
\textit{Graph influence $I_G$ on $u$ is: $I_G(u) = \Vert [I_{u,v_1}, \dots, I_{u,v_n}] \Vert_1$, where $[I_{u,v_1}, \dots, I_{u,v_n}]$ is a vector representing the influence of other nodes on $u$. }

The graph influence of graph $G$ on node $u$ is: $I_G(u) = \sum_{i \in \mathcal{V}} I_{u,v_i}$. Similarly, the influence of a $h$-hop neighborhood subgraph $S_u$ on node $u$ is: $I_{S_u}(u) =  \sum_{i \in \mathcal{V}^u} I_{u,v_i}$ where $\mathcal{V}^u$ contains nodes that are at most $h$ hops away from node $u$, \ie, $\{i_x | d(i_x, u) \le h\}$. 

\textbf{Definition (Graph Influence Loss)}:
\textit{Graph influence loss $R_h$ is defined as: $R_h(u) = I_G(u) - I_{S_u}(u)$, where $I_G(u)$ is the influence of entire graph $G$ and $I_{S_u}(u)$ is the influence of subgraph $S_u$.}

Notationally, we denote $m$ paths between node $u$ and $v$ as: $p^1, ..., p^m$. We use $p^x_i, p^x_j, ..., p^x_{n_v}$ to represent nodes occurring on path $p^x$ and use $n_v$ to denote the length of the path. 

\subsection{Theorem and Proof}

\textbf{Theorem 1~(Decaying Property of Node Influence).} \textit{Let $t$ be a path between node $u$ and node $v$ and let $D_\mathrm{GM}^t$ be a geometric mean of node degrees occurring on path $t$. Let $D^{t_*}_\mathrm{GM} = \min_t \{D_\mathrm{GM}^t\}$ and $h_* = d(u,v)$. Consider the node influence $I_{u,v}$ from $v$ to $u$. Then, $I_{u,v} \le C / (D^{t_*}_{\mathrm{GM}})^{h_*}. $}

\begin{proof}

Using the GCN propagation rule, we see that node $u$'s output is defined as:
\begin{equation*}
    \mathbf{x}_u^{(\infty)} = \frac{1}{D_{uu}} \sum_{i \in \mathcal{N}(u)} a_{u i} \mathbf{x}_i^{(\infty)},
\end{equation*}
where $a_{ui}$ is the edge weight between node $u$ and node $i$. In our datasets and many other real-world graphs, all edges have the same weight 1. 

By an expansion of nodes in the neighbor $\mathcal{N}(u)$, we have: 
\begin{equation*}
    \mathbf{x}_u^{(\infty)} = \frac{1}{D_{uu}} \sum_{i \in \mathcal{N}(u)} a_{ui} \frac{1}{D_{ii}} \sum_{j \in \mathcal{N}(i)} a_{ij} \mathbf{x}_j^{(\infty)}.
\end{equation*}

In the same logic, it can be further expanded as:
\begin{equation} \label{eq:expansion}
    \mathbf{x}_u^{(\infty)} = \frac{1}{D_{uu}} \sum_{i \in \mathcal{N}(u)} a_{ui} \frac{1}{D_{ii}} \sum_{j \in \mathcal{N}(i)} a_{ij} \cdot \cdot \cdot \frac{1}{D_{mm}} \sum_{o \in \mathcal{N}(m)} a_{mo} \mathbf{x}_o^{(\infty)}.
\end{equation}

The node influence $I_{u,v} = \Vert \frac{\partial \mathbf{x}^{(\infty)}_u}{\partial \mathbf{x}^{(\infty)}_v} \Vert$ can then be computed via:

\begin{align*}
    \Vert \frac{\partial \mathbf{x}^{(\infty)}_u}{\partial \mathbf{x}^{(\infty)}_v} \Vert & = \Vert \frac{\partial}{\partial \mathbf{x}^{(\infty)}_v} \left(\frac{1}{D_{u u}} \sum_{i \in \mathcal{N}(u)} a_{u i} \frac{1}{D_{ii}} \sum_{j \in \mathcal{N}(i)} a_{ij} \cdot \cdot \cdot \frac{1}{D_{mm}} \sum_{o \in \mathcal{N}(m)} a_{mo} \mathbf{x}_o^{(\infty)}\right) \Vert & [1] \\
    & = \Vert \frac{\partial}{\partial \mathbf{x}^{(\infty)}_v} \Biggl( \left( \frac{1}{D_{u u}} a_{(u p^1_i)} \frac{1}{D_{p^1_i p^1_i}} a_{(p^1_i p^1_j)}\cdot \cdot \cdot \frac{1}{D_{p^1_{n_1} p^1_{n_1}}} a_{(p^1_{n_1} v)} \mathbf{x}^{(\infty)}_v \right) &  \\
    & + \cdot \cdot \cdot + \left( \frac{1}{D_{u u}} a_{(u p^m_i)} \frac{1}{D_{p^m_i p^m_i}} a_{(p^m_i p^m_j)} \cdot \cdot \cdot \frac{1}{D_{p^m_{n_m} p^m_{n_m}}} a_{(p^m_{n_m} v)} \mathbf{x}^{(\infty)}_v \right) \Biggr) \Vert & [2] \\
    & =  \Vert \frac{\partial \mathbf{x}^{(\infty)}_v}{\partial \mathbf{x}^{(\infty)}_v} \Vert \cdot \vert \left( \frac{1}{D_{u u}} a_{(u p^1_i)} \frac{1}{D_{p^1_i p^1_j}} a_{(p^1_i p^1_j)}\cdot \cdot \cdot \frac{1}{D_{p^1_{n_1} p^1_{n_1}}} a_{(p^1_{n_1} v)} \right) \\
    & + \cdot \cdot \cdot + \left( \frac{1}{D_{u u}} a_{(u p^m_i)} \frac{1}{D_{p^m_i p^m_i}} a_{(p^m_i p^m_j)} \cdot \cdot \cdot \frac{1}{D_{p^m_{n_m} p^m_{n_m}}} a_{(p^m_{n_m} v)} \right) \vert & [3]\\
    & = \vert \left( \frac{1}{D_{u u}D_{p^1_i p^1_i}\cdot \cdot \cdot D_{p^1_{n_1} p^1_{n_1}}} a_{(u p^1_i)}  a_{(p^1_i p^1_j)} \cdot \cdot \cdot a_{(p^1_{n_1} v)}  \right)  \\
    & + \cdot \cdot \cdot + \left( \frac{1}{D_{u u}D_{p^1_i p^1_j}\cdot \cdot \cdot D_{p^m_{n_m} p^m_{n_m}}} a_{(u p^m_i)}  a_{(p^m_i p^m_j)} \cdot \cdot \cdot a_{(p^m_{n_m} v)} \right) \vert & [4]\\ 
    &\le \vert m * \mathrm{max} \Biggl(\left( \frac{1}{D_{u u}D_{p^1_i p^1_i}\cdot \cdot \cdot D_{p^1_{n_1} p^1_{n_1}}} a_{(u p^1_i)}  a_{(p^1_i p^1_j)} \cdot \cdot \cdot a_{(p^1_{n_1} v)}  \right) \\
    &, \cdot \cdot \cdot, \left( \frac{1}{D_{u u}D_{p^1_i p^1_i}\cdot \cdot \cdot D_{p^m_{n_m} p^m_{n_m}}} a_{(u p^m_i)}  a_{(p^m_i p^m_j)} \cdot \cdot \cdot a_{(p^m_{n_m} v)} \right) \Biggr) \vert & [5] \\
    & = \vert  m * \left(\frac{1}{D_{u u}D_{p^{t_*}_i p^{t_*}_i}\cdot \cdot \cdot D_{p^{t_*}_{n_*} p^{t_*}_{n_*}}} a_{(u p^{t_*}_i)}  a_{(p^{t_*}_i p^{t_*}_j)} \cdot \cdot \cdot a_{(p^{t_*}_{n_m} v)} \right) \vert & [6]\\
    & = \vert  C * \left(\frac{1}{ \sqrt[{n_*}]{ D_{u u}D_{p^{t_*}_i p^{t_*}_i}\cdot \cdot \cdot D_{p^{t_*}_{n_*} p^{t_*}_{n_*}}}} \right)^{n_*}  \vert & [7]\\
    & =  C / (D^{t_*}_{\mathrm{GM}})^{n_*}  & [8] \\
    & \le C / (D^{t_*}_{\mathrm{GM}})^{h_*}. & [9]
\end{align*}
\end{proof}

\xhdr{Explanations of the proof} (Line 1) Substitution of the term $\mathbf{x}^{(\infty)}_u$ via Equation~\ref{eq:expansion}. (Line 2) Since we are calculating the gradient between two feature vectors $\mathbf{x}^{(\infty)}_v$ and $\mathbf{x}^{(\infty)}_u$, the partial derivative on nodes that are not in the paths $p^1, ..., p^m$ between node $u$ and $v$ becomes 0 and these nodes are thus removed. Note that each term in line 2 is for one path between node $u$ and $v$. This also means that the feature influence can be decomposed into the sum of all paths feature influence. (Line 3) We separate the scalar terms and the derivative term within the matrix norm and then uses the absolute homogeneous property (\ie $\Vert \alpha \mathbf{A} \Vert = \vert \alpha \vert \Vert \mathbf{A} \Vert $) of the matrix norm to convert them into the matrix norm of Jacobian matrix times the absolute value of the scalar. (Line 4) We know the Jacobian of the same vectors $\frac{\partial \mathbf{x}^{(\infty)}_v}{\partial \mathbf{x}^{(\infty)}_v}$ is the identity matrix $\mathbf{I}$ and we know for any subordinate norm ($\Vert \mathbf{A} \Vert = \text{sup}_{\Vert \mathbf{x} \Vert = 1} \{ \Vert \mathbf{Ax} \Vert \}$), we have $\Vert \mathbf{I} \Vert = 1$. Therefore,  $\Vert \frac{\partial \mathbf{x}^{(\infty)}_v}{\partial \mathbf{x}^{(\infty)}_v} \Vert = 1$. We also move the degree and edge weight terms around to group each together for each path. (Line 5) Notice now the equation becomes the sum of terms around paths between node $u$ and $v$. We can identify the maximum of these terms and it is smaller than the $m$ times the maximum term. We denote the term with the maximum value as the path $p^{t_*}$. (Line 6) Clean the notations. (Line 7) We move all the constant as $C$ in the last line. Note that in all of our datasets and many real-world datasets, the  network is usually binary and the edge weight $a$ are non-negative (in many cases $a = 1$), thus we remove the absolute values. (Line 8) We rephrase the degree term into geometric mean format $1/D_{GM}$. (Line 9) $h_*$ is the shortest path and shorter than $n_*$, thus the inequality. We then obtain the bound of the node influence.

Note that if we assume degree of nodes along paths are random, then the $p^{t_*}$, which is the path that has the smallest geometric mean of node degrees, is the shortest path from node $u$ to $v$.

\section{Theorem 2 and its Proof} \label{appendix:2}

\textbf{Theorem 2~(Local Subgraph Preservation Property).} \textit{Let $S_u$ be a local subgraph for node $u$ with neighborhood size $h$. Let node $v$ be defined as: $v = \mathrm{argmax}_w(\{I_{u,w} | w \in \mathcal{V} \setminus \mathcal{V}^u\})$. Let $\bar{t}$ be a path between $u$ and $v$ and let $D_{\mathrm{GM}}^{\bar{t}}$ be a geometric mean of node degrees occurring on path $\bar{t}$. Let $D^{\bar{t}_*}_\mathrm{GM} = \min_{\bar{t}}\{D_\mathrm{GM}^{\bar{t}}\}$. The following holds: $R_h(u) \le C / (D^{\bar{t}_*}_\mathrm{GM})^{h+1}.$}

\begin{proof}
\begin{align*}
        R_h(u) & = I_{G}(u) - I_{S_u}(u) & [1]\\
        & =  \left( \Vert \frac{\partial \mathbf{x}^{(\infty)}_u}{\partial \mathbf{x}^{(\infty)}_1} \Vert + \cdot \cdot \cdot + \Vert \frac{\partial \mathbf{x}^{(\infty)}_u}{\partial \mathbf{x}^{(\infty)}_{n}} \Vert \right) - \left( \Vert \frac{\partial \mathbf{x}^{(\infty)}_u}{\partial \mathbf{x}^{(\infty)}_{i_1}} \Vert + \cdot \cdot \cdot + \Vert \frac{\partial \mathbf{x}^{(\infty)}_u}{\partial \mathbf{x}^{(\infty)}_{i_m}} \Vert \right)& [2] \\
        &= \Vert \frac{\partial \mathbf{x}_u}{\partial \mathbf{x}_{t_1}} \Vert + \Vert \frac{\partial \mathbf{x}_u}{\partial \mathbf{x}_{t_2}} \Vert + \cdot \cdot \cdot + \Vert \frac{\partial \mathbf{x}_u}{\partial \mathbf{x}_{t_{n-m}}} \Vert & [3]\\ 
        & \le C_{t_1} / (D^{t_1}_{\mathrm{GM}})^{h_{t_1}}  + \cdot \cdot \cdot +  C_{t_{n-m}} / (D^{t_{n-m}}_{\mathrm{GM}})^{h_{t_{n-m}}}  & [4] \\
        & \le (n - m) * C_{\bar{t}_*} / (D^{\bar{t}_*}_{\mathrm{GM}})^{h_{\bar{t}_*}} & [5] \\
        &\le (n - m) * C_{\bar{t}_*} / (D^{\bar{t}_*}_{\mathrm{GM}})^{h+1} & [6]\\
        & = C / (D^{\bar{t}_*}_{\mathrm{GM}})^{h+1} & [7] 
    \end{align*}
\end{proof}

\xhdr{Explanation of the proof} (Line 1) Definition of graph influence loss. (Line 2) Substitution of definition. (Line 3) Subtract same node influence term from the subgraph and the entire network. (Line 4) Use Theorem 1. (Line 5) They are smaller than $(n-m)$ times the maximum term, which is the node that has the highest node influence in the node set that is outside of the immediate neighborhood of $u$. (Line 6) We know nodes in the outside of the immediate neighborhood of $u$ is more than $h$ hops away from the node $u$, hence the path length $h_{t_*}$ between the maximum influence node to the node $u$ is larger than $h$. (Line 7) Move the constant term and we have the result.

\section{Algorithm Overview} \label{appendix:algo}

We provide the pseudo-code for \name in Algorithm~\ref{algo1}.

\begin{algorithm}[h]
\SetAlgoLined
\DontPrintSemicolon
\textbf{Input}: Graphs $\mathcal{G} = \{G_1, ..., G_N\}$; Randomly initialized $\theta: [\theta_\text{GNN}]$. \;
$S_1, S_2, ..., S_n = \text{Subgraph}(\mathcal{G})$ via step \#1 \tcp*{Local subgraph construction}
$\{\mathcal{T}\} = \{\mathcal{T}_1, \mathcal{T}_2, ..., \mathcal{T}_m\}\sim p(\mathcal{T})$ via step \#1\tcp*{Meta-task construction} 
\While{not done}{
$\{\mathcal{T}_s\} \leftarrow \text{sample}(\{\mathcal{T}\})$ \tcp*{Sample a batch of tasks}
\For{$\mathcal{T}_i \in \{\mathcal{T}_s\}$}{
$(\{S\}_\text{support},\mathbf{y}_\text{support}) \leftarrow \mathcal{T}^\text{support}_i$ \tcp*{Mini-batching support subgraphs}
$(\{S\}_\text{query},\mathbf{y}_\text{query}) \leftarrow \mathcal{T}^\text{query}_i$ \tcp*{Mini-batching query subgraphs}
$\theta_0 = \theta$ \;
\For(\tcp*[f]{Update step $j$}){j in 1, ..., $\eta$}{
    $\mathbf{H}_\text{support} \leftarrow \mathrm{GNN}_{\theta_{j-1}}(\{S\}_\text{support})_\text{centroid}$ via step \#2 and \#3\;
    $\mathbf{C} = \frac{1}{N_{k_\text{support}}} \sum (\mathbf{H}_\text{support})$ via step \#4\tcp*{Support prototypes}
    $\mathbf{p} = \frac{\mathrm{exp}\left( - \Vert \mathbf{H}_\text{support}  - \mathbf{C} \Vert \right)}{\sum_{\textsc{Y}_i} \mathrm{exp}\left( - \Vert \mathbf{H}_\text{support}  - \mathbf{C} \Vert \right)} $ via step \#5  \;
    $\mathcal{L}_\text{support} = \mathrm{L}(\mathbf{p}, \mathbf{y}_\text{support})$ via step \#5 \;
    $\theta_j = \theta_{j-1} - \alpha \nabla \mathcal{L}_\text{support}$ via step \#6 \tcp*{Inner loop update}
    $\mathbf{H}_\text{query} \leftarrow \mathrm{GNN}_{\theta_j}(\{S\}_\text{query})_\text{centroid}$ via step \#7 and \#8\;
   $\mathbf{p} = \frac{\mathrm{exp}\left( - \Vert \mathbf{H}_\text{query}  - \mathbf{C} \Vert \right)}{\sum_{\textsc{Y}_i} \mathrm{exp}\left( - \Vert \mathbf{H}_\text{query}  - \mathbf{C} \Vert \right)} $ via step \#9 \;
    $\mathcal{L}_\text{query}^{ij} \leftarrow  \mathrm{L}(\mathbf{p},\mathbf{y}_\text{query})$ via step \#9
    }
    }
    $\theta = \theta - \beta \nabla \sum_{i} \mathcal{L}_\text{query}^{iu}$ via step \#10 \tcp*{Outer loop update}
}
 \caption{\name Algorithm. Steps in the algorithm correspond to steps in Figure~\ref{fig:method-overview}.}
\label{algo1}
\end{algorithm}

\section{Further Details on Datasets} \label{appendix:dataset}

We proceed by describing the construction and processing of synthetic as well as real-world datasets. \name implementation as well as all datasets and the relevant data loaders are available at \url{https://github.com/mims-harvard/G-Meta}.

\subsection{Synthetic Datasets}
We have two synthetic datasets where labels are depended on local structural roles. They are to show \mname's ability to capture local network structures. For the first synthetic dataset, we use the graphs with planted structural equivalence from GraphWave~\cite{donnat2018learning}. We use a cycle basis network and attach a distribution of shapes: House, Star, Diamond, Fan on the cycle basis. The label of each node is the structural role in different shapes. Hence, the label reflects the local structural information. We also add n random edges to add noise. For the single graph and disjoint label setting, we use 500 nodes for the cycle basis, and add 100 shapes for each type with 1,000 random edges. In the multiple graph setting, we sample 10 graphs with varying distribution of number of shapes for each graph. Each graph uses 50 nodes for the cycle basis, and add randomly generated [1-15] shapes for each type with 100 random edges. There are 17 labels. 
To model local structural information under a more realistic homophily network, we first construct a Barab{\'a}si-Albert (BA) network with 200 nodes and 3 nodes are preferential attached based on the degrees. Then, we plant shapes to the BA network by first sampling nodes and adding edges corresponding to the shapes. Then, to generate label, we compute the Graphlet Distribution Vector~\cite{prvzulj2007biological} for each node, which characterizes the local network structures and then we apply spectral clustering on this vector to generate the labels. There are 10 labels in total. For multiple graph setting, the same varying distribution of numbers of shapes as in the cycle dataset are used to plant each BA network. See Figure~\ref{fig:synthetic-data-illustration} for a visual illustration of synthetic datasets.

\subsection{Real-world Datasets and Novel Graph Meta-Learning Datasets}
We use three real world datasets for node classification and two real world datasets for link prediction to evaluate \name. (1) arXiv is a citation network from the entire Computer Science arXiv papers, where features are title and abstract word embeddings, and labels are the subject areas~\cite{hu2020open}. (2) Tissue-PPI is 24 protein-protein interaction networks from different tissues, where features are gene signatures and labels are gene ontology functions~\cite{zitnik2017predicting,hamilton2017inductive}. Each label is a binary protein function classification task. We select the top 10 balanced tasks. (3) Fold-PPI is a novel dataset, which we constructed for the multiple graph and disjoint label setting. It has 144 tissue networks~\cite{zitnik2017predicting}, and the labels are classified using protein structures defined in SCOP database~\cite{andreeva2020scop}. We screen fold groups that have more than 9 unique proteins across the networks. It results in 29 unique labels. The features are conjoint triad protein descriptor~\cite{shen2007predicting}. In Fold-PPI, the majority of the nodes do not have associated labels. Note that \name operates on label scarce settings. (4) For link prediction, the first dataset FirstMM-DB~\cite{neumann2013graph} is the standard 3D point cloud data, which consists of 41 graphs. (5) The second link prediction dataset is the Tree-of-Life dataset. This is a new dataset, which we constructed based on 1,840 protein interaction networks (PPIs), each originating from a different species~\cite{zitnik2019evolution}. Node features are not provided, we use node degrees instead.

\section{Further Details on Hyperparameter Selection}\label{appendix:param}

We use random hyperparameter search over the following set of hyperparameters. For task numbers in each batch, we use 4, 8, 16, 32, 64; for inner update learning rate, 1\e{-2}, 5\e{-3}, 1\e{-3}, 5\e{-4}; for outer update learning rate, 1\e{-2}, 5\e{-3}, 1\e{-3}, 5\e{-4}; for hidden dimension, we select from 64, 128, and 256.

For arxiv-ogbn dataset, we set task numbers to 32, inner update learning rate to 1\e{-2}, outer update learning rate to 1\e{-3}, and hidden dimensionality to 256. For Tissue-PPI dataset, we set task numbers to 4, inner update learning rate to 1\e{-2}, outer update learning rate to 5\e{-3}, and hidden dimensionality to 128. For Fold-PPI dataset, we set task numbers to 16, inner update learning rate to 5\e{-3}, outer update learning rate to 1\e{-3}, and hidden dimensionality to 128. For FirstMM-DB dataset, we set task numbers to 8, inner update learning rate to 1\e{-2}, outer update learning rate to 5\e{-4}, and hidden dimensionality to 128. For Tree-of-Life dataset, we set task numbers to 8, inner update learning rate to 5\e{-3}, outer update learning rate to 5\e{-4} and hidden dimensionality to 256.

\section{Further Details on Baselines}
We use nine baselines. (1)~\underline{Meta-Graph}~\cite{bose2019meta} uses VGAE to do few-shot multi-graph link prediction. It uses a graph signature function to capture the characteristics of a graph, which enables knowledge transfer. Then, it applies MAML to learn across graphs. (2)~\underline{Meta-GNN}~\cite{zhou2019meta} applies MAML~\cite{finn2017model} to Simple Graph Convolution(SGC)~\cite{pmlr-v97-wu19e} on the single graph disjoint label problem.  (3)~Few-shot Graph Isomorphism Network (\underline{FS-GIN})~\cite{xu2018how} applies GIN on the entire graph and retrieve the few-shot nodes to propagate loss and enable learning. Similarly, (4)~Few-shot SGC (\underline{FS-SGC})~\cite{pmlr-v97-wu19e} switches GIN to SGC for GNN encoder. Note that the previous four baselines only work in a few graph meta learning problems. We also test on different meta-learning models, using the top performing ones in~\cite{triantafillou2019meta}. We operate on subgraph level for them since it allows comparison in all graph meta-learning problems.  \underline{No-Finetune} performs training on the support set and use the trained model to classify each query example, using only meta-testing set, \ie without access to the external graphs or labels to transfer.  \underline{KNN}~\cite{dudani1976distance, triantafillou2019meta} first trains a GNN using all data in the meta-training set and it is used as an embedding function. Then, during meta-testing, it uses the label of the voted K-closest example in the support set for each query example. \underline{Finetune}~\cite{triantafillou2019meta} uses the embedding function generated from meta-training set and the models are then finetuned on the meta-testing set. \underline{ProtoNet}~\cite{snell2017prototypical} applies prototypical learning on each subgraph embeddings, following the standard few-shot learning setups. \underline{MAML}~\cite{finn2017model} switches ProtoNet to MAML model as the meta-learner.

\section{Further Details on Performance Evaluation}\label{appendix:perf}

All of our experiments are done on an Intel Xeon CPU 2.50GHz and using an NVIDIA K80 GPU.

For synthetic datasets disjoint label problems, we use a 2-way setup. In the shared label problem, the cycle graph has 17 labels and the BA graph has 10 labels. The results use 5 gradient update steps in meta-training and 10 gradient update steps in meta-testing. For real-world datasets node classification uses 3-shots and link prediction uses 16-shots. For disjoint labels problems, we set it to 3-way classification task. The results use 20 gradient update steps in meta-training and 10 gradient update steps in meta-testing. For Tissue-PPI, we use the average of ten 2-way protein function tasks where each task is performed three times. For link prediction problem, to ensure the support and query set are distinct in all meta-training tasks, we separate a fixed 30\% of edges for support set and 70\% of edges for query set as a preprocessing step for every graph. 

\section{Parameter Studies}\label{appendix:param_studies}
We select Fold-PPI from node classification and FirstMM-DB from link prediction to conduct the parameter studies. We then conduct five runs on each setup and report the average performance below. For varying the number of $k$, we experiment on $k = 1, 3, 10$ for Fold-PPI and $k = 16, 32, 64$ for FirstMM-DB. We find linear trend between $k$ and predictive performance: for Fold-PPI, the performance increases from 0.403 to 0.561 to 0.663; for FirstMM-DB, \name increases from 0.758 to 0.784 to 0.795. Then, we vary the number of $h$-hops neighbor. We try $h = 1, 2, 3$. For Fold-PPI, the performance is 0.399, 0.561 and 0.427. For FirstMM-DB, the result is 0.616, 0.784 and 0.837. It seems that $h = 1$ has the worst performance but for $h = 3$, it depends on the dataset. For Fold-PPI, $h = 2$ is significantly better than $h = 3$ but it is not the case for FirstMM-DB. Since $h = 2$ has pretty close result with $h = 3$ in FirstMM-DB, we suggest to set $h = 2$ for stable performance.

\section{Further Results on Synthetic Datasets}\label{appendix:syn}

Full version is reported in Table~\ref{tab:res_syn_appendix}. The large standard deviation is because we sample only two-labels for meta-testing in one data fold, due to the limit number of labels in synthetic datasets. If the shape structural roles corresponding to the label set in the meta-testing set is structurally distinct from all the meta-training label sets, the performance would be bad since there is no transferability for meta-learner to learn. In our data splits, we find there is one fold that performs bad across all methods, thus the reason for the large standard deviation. For real-world datasets, as the label set is large, we sample more labels for meta-testing (\eg, five labels for 3-way classification where each meta-testing task is of $\binom{5}{3}$ label sets) such that the result is averaged across different label sets. This makes the standard deviation smaller. In both settings, the mean accuracy reflects the predictive performance.

\begin{table}[h]
    \centering
    \def\arraystretch{0.80}
    \caption{\textbf{Further results on graph meta-learning performance for synthetic datasets.} Five-fold average multi-class classification accuracy on synthetic datasets for 1-shot node classification in various graph meta-learning problem. N/A means the method does not work in this task setting. The disjoint label setting uses 2-way setup and in the shared labels settings, cycle basis has 17 labels and BA has 10 labels. The results use 5-gradient update steps in meta-training and 10-gradient update steps in meta-testing. See Figure~\ref{fig:synthetic-data-illustration} for a visual illustration of synthetic datasets. }
    \begin{adjustbox}{width=\textwidth}
    \begin{tabular}{l|cc|cc|cc}
    \toprule
    \rowcolor{GrayDark} Graph Meta- & \multicolumn{2}{c|}{Single graph} & \multicolumn{2}{c|}{Multiple graphs} & \multicolumn{2}{c}{Multiple graphs} \\ 
    \rowcolor{GrayDark} Learning Problem & \multicolumn{2}{c|}{Disjoint labels} & \multicolumn{2}{c|}{Shared labels} & \multicolumn{2}{c}{Disjoint labels} \\ \midrule 
     \rowcolor{GrayLight}Prediction Task &\multicolumn{2}{c|}{Node} & \multicolumn{2}{c|}{Node} &\multicolumn{2}{c}{Node} \\ \midrule
    Base graph & Cycle & BA & Cycle & BA & Cycle & BA \\ \midrule
    \name~(Ours) &0.872\std{0.129}& \textbf{0.867}\std{0.147}& \textbf{0.542}\std{0.045}&\textbf{0.734}\std{0.038}&\textbf{0.767}\std{0.178}&\textbf{0.867}\std{0.209} \\
    Meta-Graph  & N/A & N/A & N/A & N/A &  N/A &  N/A \\
    Meta-GNN & 0.720\std{0.218}&0.694\std{0.112} & N/A & N/A & N/A & N/A \\
    FS-GIN & 0.684\std{0.144} & 0.749\std{0.106} & N/A & N/A & N/A & N/A \\
    FS-SGC &0.574\std{0.092} &0.715\std{0.100} & N/A & N/A & N/A & N/A \\\midrule
    KNN &\textbf{0.918}\std{0.063} & 0.804\std{0.193} & 0.343\std{0.038} & 0.710\std{0.027} & 0.753\std{0.150}&0.769\std{0.199} \\
    No-Finetune & 0.509\std{0.011} &0.567\std{0.105}&0.059\std{0.001}& 0.265\std{0.141} &0.592\std{0.105}&0.577\std{0.104} \\
    Finetune &0.679\std{0.043}&0.671\std{0.055} & 0.385\std{0.089}& 0.517\std{0.160} &0.599\std{0.085}&0.629\std{0.055} \\
    ProtoNet &0.821\std{0.197}&0.858\std{0.144}&0.282\std{0.045}&0.657\std{0.034}&0.749\std{0.182}&0.866\std{0.212} \\
    MAML &0.842\std{0.207}&0.848\std{0.212}&0.511\std{0.050}&0.726\std{0.023}&0.653\std{0.093}&0.844\std{0.202} \\ 
    \bottomrule
    \end{tabular}
    \end{adjustbox}
    \label{tab:res_syn_appendix}
\end{table}

\end{appendices}

\end{document}